\pdfoutput=1

\documentclass[11pt]{article}

\usepackage[preprint]{acl}

\usepackage{longtable}
\usepackage{geometry}
\usepackage{graphicx}
\usepackage{adjustbox}
\usepackage{amssymb}
\usepackage{subfigure}
\usepackage{subcaption}
\usepackage{array,booktabs,ragged2e}
\usepackage{multirow}
\usepackage{multicol}
\usepackage{times}
\usepackage{latexsym}
\usepackage{arydshln}
\usepackage{hyperref}
\usepackage{pifont}
\usepackage{float}
\usepackage{bm}
\usepackage{microtype}
\usepackage{inconsolata}
\usepackage{xcolor}
\usepackage{xspace}

\usepackage{placeins}
\usepackage{afterpage}
\usepackage{amsfonts}
\usepackage{amsmath}

\usepackage[T1]{fontenc}
\usepackage[utf8]{inputenc}

\definecolor{low}{HTML}{D62F2F}
\definecolor{high}{HTML}{485694}

\newcommand{\low}{}
\newcommand{\high}{}

\newcommand{\bench}{\textbf{\textsc{besstie}}\xspace}
\newcommand{\xmark}{\ding{53}} 

\newcommand{\bert}{\textsc{bert}\xspace}
\newcommand{\roberta}{\textsc{roberta}\xspace}
\newcommand{\albert}{\textsc{albert}\xspace}
\newcommand{\mbert}{\textsc{mbert}\xspace}
\newcommand{\mdis}{\textsc{mdistil}\xspace}
\newcommand{\xlmr}{\textsc{xlm-r}\xspace}
\newcommand{\gemma}{\textsc{gemma}\xspace}
\newcommand{\mistral}{\textsc{mistral}\xspace}
\newcommand{\qwen}{\textsc{qwen}\xspace}

\newcommand{\google}{\textsc{google}\xspace}
\newcommand{\reddit}{\textsc{reddit}\xspace}

\newcommand{\f}{\textsc{F-Score}\xspace}

\newcommand{\greentext}[1]{\text{#1}}

\title{BESSTIE: A Benchmark for Sentiment and Sarcasm Classification for Varieties of English}

\author{
 \textbf{Dipankar Srirag\textsuperscript{1}}\quad
 \textbf{Aditya Joshi\textsuperscript{1}}\quad
 \textbf{Jordan Painter\textsuperscript{2}}\quad
 \textbf{Diptesh Kanojia\textsuperscript{2}}
 \\
    \textsuperscript{1}University of New South Wales, Sydney, Australia
\\
    \textsuperscript{2}Institute for People-Centered AI, University of Surrey, Surrey, United Kingdom
\\
    \texttt{\{d.srirag,aditya.joshi\}@unsw.edu.au}\quad\texttt{\{jp1106,d.kanojia\}@surrey.ac.uk}
 }

\begin{document}
\maketitle
\begin{abstract}
Despite large language models (LLMs) being known to exhibit bias against non-standard language varieties, there are no known labelled datasets for sentiment analysis of English. To address this gap, we introduce \bench, a benchmark for sentiment and sarcasm classification for three varieties of English: Australian (en-AU), Indian (en-IN), and British (en-UK). We collect datasets for these \greentext{language varieties} using two methods: location-based for Google Places reviews, and topic-based filtering for Reddit comments. To assess whether the dataset accurately represents these varieties, we conduct two validation steps: (a) manual annotation of language varieties and (b) automatic language variety prediction. Native speakers of the language varieties manually annotate the datasets with sentiment and sarcasm labels. We perform an additional annotation exercise to validate the reliance of the annotated labels. Subsequently, we fine-tune \greentext{nine} large language models (LLMs) (representing a range of encoder/decoder and mono/multilingual models) on these datasets, and evaluate their performance on the two tasks. Our results show that the models consistently perform better on inner-circle varieties \greentext{(\textit{i.e.}, en-AU and en-UK)}, in comparison with en-IN, particularly for sarcasm classification. We also report challenges in cross-variety generalisation, highlighting the need for language variety-specific datasets such as ours. \bench promises to be a useful evaluative benchmark for future research in equitable LLMs, specifically in terms of language varieties. The \bench dataset is publicly available at: \url{https://huggingface.co/datasets/unswnlporg/BESSTIE}. 
\end{abstract}

\section{Introduction}
Benchmark-based evaluation~\cite{socher-etal-2013-recursive} of large language models (LLMs) is the prevalent norm in natural language processing (NLP). Benchmarks provide labelled datasets~\cite{NEURIPS2019_4496bf24,muennighoff2023mtebmassivetextembedding} for specific tasks so that LLMs can be evaluated against them. However, most NLP benchmarks do not contain text in language varieties such as national varieties, dialects, sociolects or creoles~\cite{plank-2022-problem,https://doi.org/10.48550/arxiv.2310.19567, joshi2024naturallanguageprocessingdialects}. 
\begin{table}[t!]
    \begin{adjustbox}{width=\linewidth,center}
        \begin{tabular}{cccccc}
            \toprule
            Benchmark & Sent. & Sarc. & Eng. & Var. \\\midrule[\heavyrulewidth]
            \citet{cieliebak-etal-2017-twitter} & \checkmark & \xmark & \xmark & \xmark\\
            \citet{wang-etal-2018-glue} & \checkmark & \xmark & \checkmark & \xmark\\
            \citet{alharbi2020asad} & \checkmark & \xmark & \xmark & \checkmark\\
            \citet{abu-farha-etal-2021-overview} & \checkmark & \checkmark & \xmark & \checkmark\\
            \citet{elmadany-etal-2023-orca} & \checkmark & \checkmark & \xmark & \checkmark\\
            \citet{faisal-etal-2024-dialectbench} & \checkmark & \xmark & \xmark & \checkmark\\\hdashline
            \bench & \checkmark & \checkmark & \checkmark & \checkmark\\
            \bottomrule
        \end{tabular}
    \end{adjustbox}
    \caption{Comparison of \bench with past benchmarks for sentiment or sarcasm classification. `Sent.' indicates sentiment classification, `Sarc.' denotes sarcasm classification, {`Eng.' denotes English, and `Var.' denotes language varieties}. A checkmark (\checkmark) denotes the availability of a particular feature, while a cross (\xmark) indicates its absence.}
    \label{tab:compare}
\end{table}
\begin{figure*}[t!]
    \centering
    \includegraphics[width=0.7\linewidth]{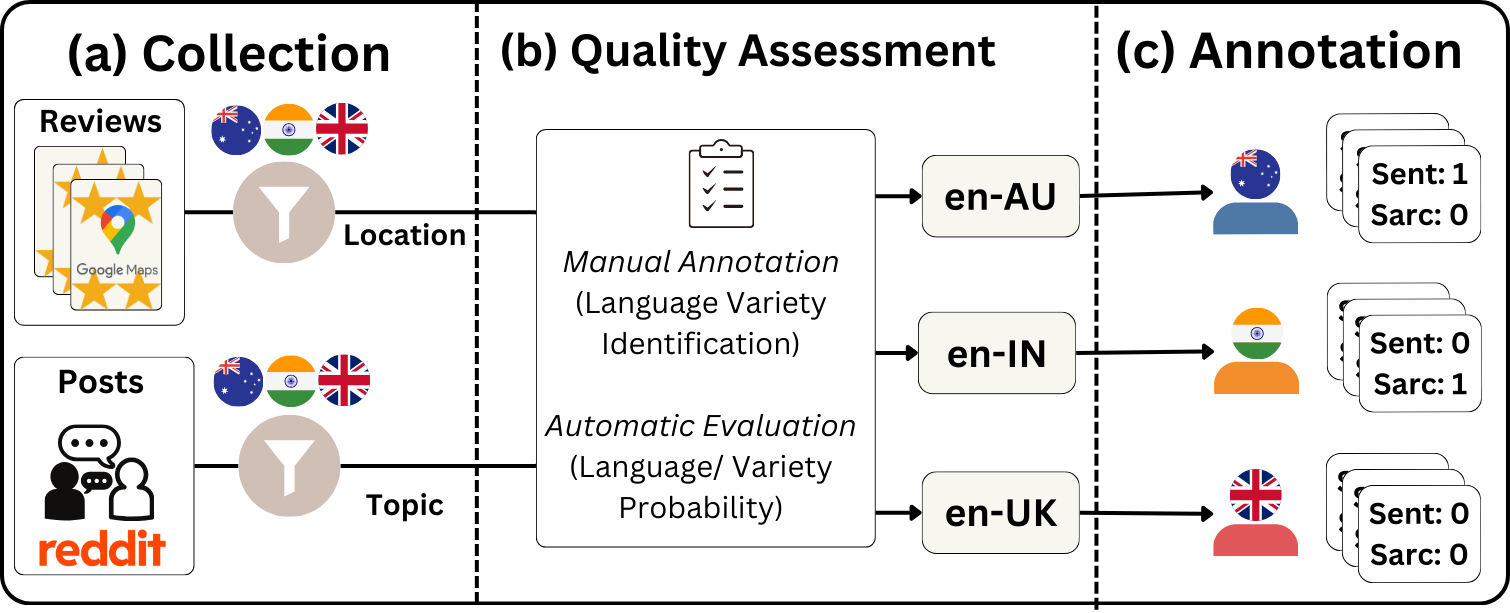}
    \caption{Creating dataset for \bench; Collection in Section~\ref{sec:curate}, Quality Assessment in Section~\ref{sec:qualass}, Annotation in~\ref{sec:annot}.}
    \label{fig:bench}
\end{figure*}
As a result, despite the superlative performance of LLMs on Standard American English, LLMs are not evaluated on other varieties of English. Knowing that LLMs can be biased against certain varieties of English, as shown by~\citet{deas-etal-2023-evaluation} and~\citet{srirag2024evaluatingdialectrobustnesslanguage} for African-American English and Indian English respectively, little empirical evidence is found for their bias towards several other varieties. An exception is Multi-VALUE~\cite{ziems-etal-2023-multi} which creates synthetic datasets of language varieties using linguistically-informed syntactic transformations and reports a degraded performance on the GLUE benchmark. However, such synthetic texts do not accurately represent the spoken or written forms of the language variety because language varieties are more than syntax and encompass orthography, vocabulary, and cultural pragmatics~\cite{nguyen202110}. As a result, labelled datasets of natural text are crucial to measure the bias in LLMs towards these varieties.

Acknowledging the role of the sentiment classification task~\cite{socher-etal-2013-recursive} in GLUE, we introduce \bench, A \textbf{BE}nchmark for \textbf{S}entiment and \textbf{S}arcasm classification for varie\textbf{TI}es of \textbf{E}nglish, namely, Australian (en-AU), Indian (en-IN), and British (en-UK). The relationship between sentiment and sarcasm classification as related tasks is well-understood in the context of sentiment analysis~\cite{chauhan2020sentiment}. \bench comprises a manually labelled dataset of textual posts along with performance results of fine-tuned LLMs on the two classification tasks. The dataset consists of text samples collected from \textit{two} web-based domains, namely, Google Places reviews, and Reddit comments, using \textit{two} filtering methods: location-based and topic-based filtering respectively. We validate the representation of these varieties through a two-step process: (a) a manual annotation exercise and (b) automatic variety identification using models fine-tuned on the ICE-Corpora~\cite{https://doi.org/10.1111/j.1467-971X.1996.tb00088.x}. Following that, we obtain manual annotations for boolean sentiment and sarcasm labels from native speakers of the language varieties. 

We then use the dataset to model both the tasks as \textit{binary classification} problems. We evaluate {nine fine-tuned} LLMs, spanning encoder/decoder and monolingual/multilingual models. Our results indicate that current models exhibit a significant decrease in performance when processing text written in en-IN, an outer-circle variety, in comparison with inner-circle varieties of English\footnote{Inner-circle English varieties are those spoken by people who use English as a first language. Outer-circle varieties are those spoken by bi/multilingual speakers.} (en-AU and en-UK). The performance on sarcasm classification is consistently low across the three varieties, providing pointers for future work. Following that, we examine the aggregate performance for the three varieties, the two tasks, and model attributes (encoder/decoder and monolingual/multilingual). We also examine cross-variety performance on all models highlighting the need for \bench. \bench evaluation provides a solid baseline to examine biases of contemporary LLMs, and will help to develop new methods to mitigate these biases. The novelty of the \bench dataset can be seen in Table~\ref{tab:compare}. \bench improves not only upon popular sentiment/sarcasm-labelled datasets but also a recent dialectal benchmark, DialectBench~\cite{faisal-etal-2024-dialectbench}, which covers sentiment classification but for dialects of languages other than English. Therefore, the \bench dataset is novel in the introduction of new language varieties (specifically, varieties of English) for both sentiment and sarcasm classification. 

The contributions of this work are: (a) Creation of a manually annotated dataset of user-generated content with sentiment and sarcasm labels for three varieties of English, called \bench; (b) Evaluation of {nine} LLMs on \bench and analysing their performance {to identify challenges arising due to language varieties}.

\section{Dataset Creation}\label{sec:method}
Figure~\ref{fig:bench} shows the process followed to create a manually annotated dataset for sentiment and sarcasm classification.
\subsection{Data {Collection}}\label{sec:curate}
We collect textual posts from two domains: Google Places reviews and Reddit comments, using two filtering methods: location-based and topic-based respectively. Location-based filtering implies that we select reviews that were posted for locations in the three countries. Specifically, we collect reviews posted in cities in AU, IN and UK, and their corresponding ratings ($1$ to $5$ \textit{stars}, {where $5$ is the highest.}) using the Google Places API\footnote{\url{https://developers.google.com/maps/documentation/places/web-service/overview}; Accessed on 10 February 2025} from the place types, also defined by the API. The criteria for city selection are based on population thresholds specific to each country (en-AU: $20$K; en-IN:$100$K; en-UK: $50$K). We filter out any non-English content using language probabilities calculated by fastText~\cite{grave2018learning} word vectors, using a threshold of $0.98$. Additionally, to mitigate the risk of including reviews written by tourists rather than residents, we exclude reviews from locations designated as \textit{`tourist attractions'} by the Google Places API.

\begin{table}[h!]
    \begin{adjustbox}{width=0.6\linewidth, center}
        \begin{tabular}{ccc}
        \toprule
            Variety & \textsc{naive} & \textsc{ours}\\\midrule[\heavyrulewidth]
            en-AU & 0.97 & 0.79\\
            en-IN & 0.94 & 0.76\\
            en-UK & 0.96 & 0.81\\\hdashline
            $\mu$ & 0.96 & 0.79\\
        \bottomrule    
        \end{tabular}
    \end{adjustbox}
    \caption{Performance of \textsc{distil} on sentiment classification with different labelling schemes. \textsc{naive} refers to labels based on 1 and 5 \textit{stars}, while \textsc{ours} corresponds to labels based on 2 and 4 \textit{stars}. $\mu$ denotes the average \f across all varieties}
    \label{tab:label}
\end{table}

We also conduct a preliminary investigation using DistilBERT-Base (\textsc{distil};\citealp{sanh2020distilbertdistilledversionbert}) to examine the influence of label semantics, the \textit{star} ratings, on the performance of sentiment classification. We experiment with \textsc{distil} on the said task using the collected Google Places reviews under two distinct labelling schemes: (a) extreme ratings, i.e., 1 and 5 \textit{stars} (\textsc{naive}); and (b) moderate ratings, i.e., 2 and 4 \textit{stars} (\textsc{ours}). Table~\ref{tab:label} presents the performance of \textsc{distil} for each variety under both schemes, reported using the macro-averaged \f. We observe that \textsc{distil} yields strong performance when trained on \textsc{naive} labels, achieving an average $\mu$ of 0.96 across the three varieties. However, task performance drops considerably with the use of the more ambiguous \textsc{ours} labels, highlighting that they introduce increased difficulty and nuanced content. Following these findings, we filter the dataset to retain only reviews with intermediate ratings (2 and 4 \textit{stars}), to better capture this nuance and avoid model overfitting to well-distinguished and  polarised examples.

To create the \reddit subset, we employ topic-based filtering and choose up to four subreddits per variety (en-AU: \textit{`melbourne', `AustralianPolitics',`AskAnAustralian'}; en-IN: \textit{`India', `IndiaSpeaks', `BollyBlindsNGossip'}; en-UK: \textit{`England', `Britain', `UnitedKingdom', `GreatBritishMemes'}), where the topics are determined by native speakers of these language varieties. We select these subreddits based on the understanding that they feature discussions specific to a variety, making it highly likely that the post authors use the corresponding language variety. For each variety, we scrape $12,000$ comments evenly across the selected subreddits, capping at $20$ comments per post and focusing on recent posts. These are then randomly sampled and standardised to $3,000$ comments per variety before manual annotation. We discard information such as user identifiers and post identifiers to maintain the anonymity of the user.

\begin{figure}[t!]
    \centering
    \includegraphics[width=0.8\linewidth]{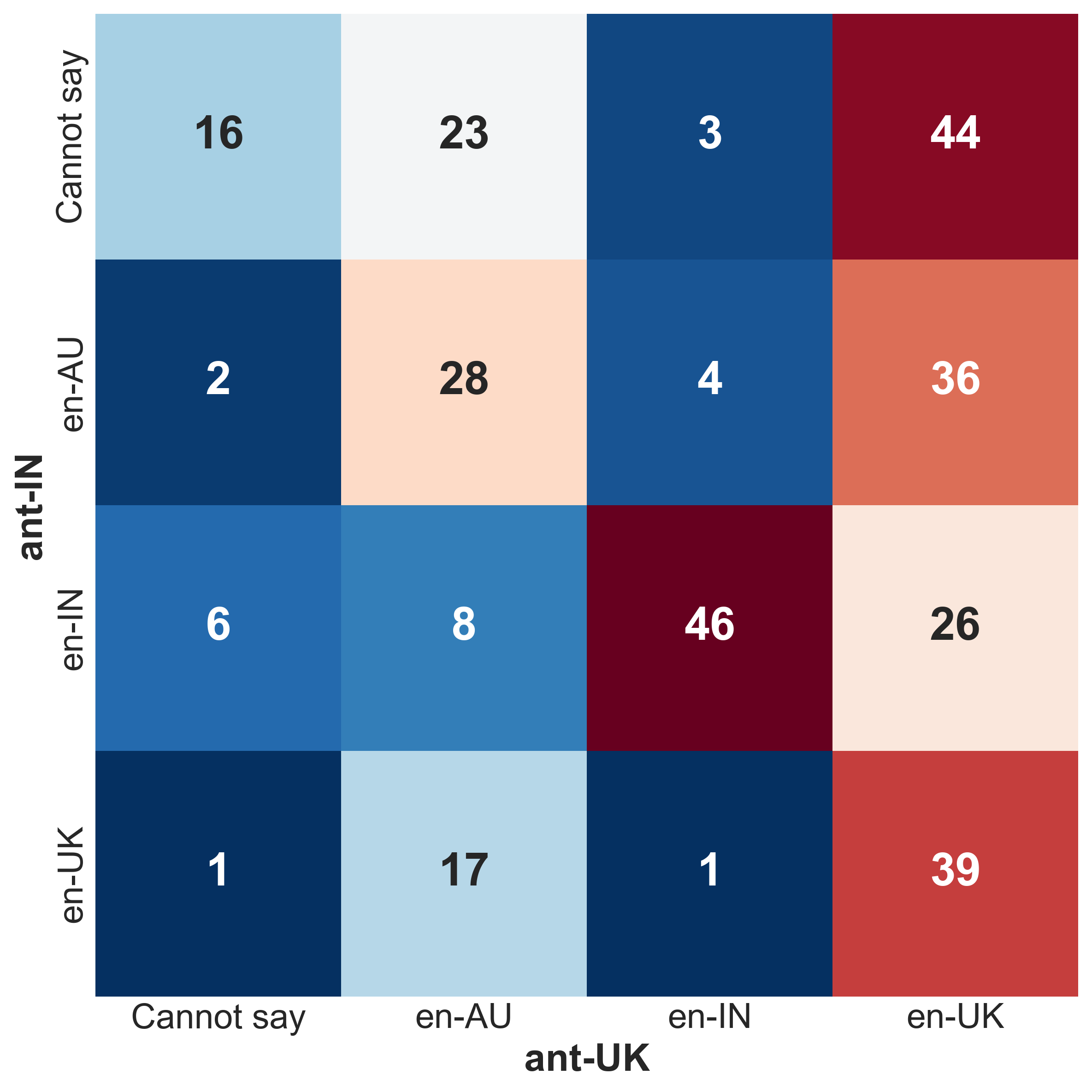}
    \caption{Confusion matrix showing the overlap in variety annotations between annotators, ant-IN, and ant-UK. Rows represent the labels assigned by ant-IN, and columns represent the labels assigned by ant-UK. The principle diagonal elements indicate agreement between annotators, while off-diagonal elements highlight disagreements.}
    \label{fig:annot}
\end{figure}
\subsection{Quality Assessment}
\label{sec:qualass}
While previous studies utilise location-based filtering~\cite{blodgett-etal-2016-demographic,sun-etal-2023-dialect}, we examine if the collected text from the two methods is indeed from these language varieties. To address this, we conduct two evaluations: (a) Manual annotation for national variety identification, and (b) Automated validation of national varieties.

\paragraph{Manual annotation} We request en-IN and en-UK annotators, asking them to manually identify the variety of a given text. Specifically, they label a random sample of $300$ texts ($150$ from reviews and $150$ from comments) as \textit{en-AU}, \textit{en-IN}, \textit{en-UK}, or \textit{Cannot say}. Annotator agreement is measured using Cohen’s kappa ($\kappa$).

With respect to the true label (based on the location or topic), results show higher agreement with the en-IN annotator ($\kappa$ = 0.41) compared to the en-UK annotator ($\kappa$ = 0.34), indicating fair to moderate consistency. The inter-annotator agreement itself is $0.26$. Figure~\ref{fig:annot} shows that annotators find it difficult to agree on an inner-circle variety, and have the highest agreement when identifying the en-IN variety, with 46 agreements, demonstrating that the subset reliably represents this variety. As a result, the two subsets (reviews and comments) \textit{collectively} form a good representative sample for language varieties.
\begin{table}[t!]
    \begin{adjustbox}{width=\linewidth, center}
        \begin{tabular}{ccccc}
        \toprule
            Variety & Subset & $P(\bar{\texttt{eng}})$ & $P(\bar{\texttt{v}})$ & \f\\\midrule[\heavyrulewidth]
            \multirow{2}{2.8em}{en-AU}
            & \google & 0.99 & 0.99 & 0.99\\\cmidrule(lr){2-5}
            & \reddit & 0.98 & 0.95 & 0.93\\\midrule[\heavyrulewidth]
            \multirow{2}{2.8em}{en-IN}
            & \google & 0.99 & 0.94 & 0.91\\\cmidrule(lr){2-5}
            & \reddit & 0.87 & 0.78 & 0.69\\\midrule[\heavyrulewidth]
            \multirow{2}{2.8em}{en-UK}
            & \google & 0.99 & 0.99 & 0.99\\\cmidrule(lr){2-5}
            & \reddit & 0.98 & 0.93 & 0.90\\
        \bottomrule    
        \end{tabular}
    \end{adjustbox}
    \caption{Data quality for each variety and subset. $P(\bar{\texttt{eng}})$ is the average language probability (calculated using fastText~\cite{grave2018learning} word vectors) of a review or comment being in English. $P(\bar{\texttt{v}})$ represents the average variety probability of a review or comment being in the corresponding variety. \f measures the performance of the variety predictor.}
    \label{tab:quality}
\end{table}
\paragraph{Automated validation} We use two predictors, a language predictor, and a variety predictor, to perform an automatic evaluation of our dataset. For language predictor, we use fastText word vectors and extract the probability of the text being in English. We then fine-tune \textsc{distil} on ICE-Corpora, to use as a variety predictor. We model the task as binary classification (i.e., inner-circle vs. outer-circle variety classification). We use the ICE-Australia~\cite{Smith2023} and ICE-India subsets of the corpus\footnote{Due to lack of access to ICE-GB, we do not include en-UK variety in training the predictor.}, focusing on the transcriptions from unprompted monologues and private dialogues (S1A and S2A headers). Similar to language predictor, we extract the probability of text being in the corresponding variety. 

$P(\bar{\texttt{eng}})$ in Table~\ref{tab:quality} is the probability that the text is written in English. $P(\bar{\texttt{v}})$ is the average variety probability. High values indicate that the variety predictor is capable of discerning text written in inner-circle varieties from those written in the outer-circle variety (i.e., en-IN). The lower language and variety probabilities on the \reddit subset from en-IN, with 0.87 and 0.78 respectively, can be attributed to the code-mixed text in the subset. These results along with our manual annotation exercise show that the collected data represents the three language varieties. The probabilities from the model are in line with the ground truth, as measured using the \f.
\begin{table}[h!]
    \begin{adjustbox}{width=0.6\linewidth,center}
    \begin{tabular}{ccc}
    \toprule
         Variety & Sent. & Sarc.\\\midrule[\heavyrulewidth]
         en-AU & 0.61 & 0.47\\
         en-IN & 0.65 & 0.51\\
         en-UK & 0.79 & 0.63\\
         \bottomrule
    \end{tabular}
    \end{adjustbox}
    \caption{Inter-annotator agreement for the validation annotation exercise. Agreement is measured between the original annotator and an independent annotator for each language variety. Here Sent. and Sarc. denote sentiment and sarcasm labels respectively.}
    \label{tab:annot}
\end{table}

\begin{table*}[t!]
    \begin{adjustbox}{width=0.85\linewidth,center}
        \begin{tabular}{cccccccc}
            \toprule
            Variety & {Subset} & Train & Valid & Test & \centering \% Pos. Sent. & \centering \% Pos. Sarc. & Avg. no. of words\\\midrule[\heavyrulewidth]
            \multirow{2}{2.8em}{en-AU}
            & \google & 946 & 130 & 270 & \centering 73\% & \centering 7\%  & 63.97\\\cmidrule(lr){2-8}
            & \reddit & 1763 & 241 & 501 & \centering 32\% & \centering 42\%  & 51.72\\\midrule[\heavyrulewidth]
            \multirow{2}{2.8em}{en-IN}
            & \google & 1648 & 225 & 469 & \centering 75\% & \centering 1\%  & 44.34\\\cmidrule(lr){2-8}
            & \reddit & 1686 & 230 & 479 & \centering 25\% & \centering 13\%  & 26.92\\\midrule[\heavyrulewidth]
            \multirow{2}{2.8em}{en-UK}
            & \google & 1817 & 248 & 517 & \centering 75\% & \centering 0\%  & 72.21\\\cmidrule(lr){2-8}
            & \reddit & 1007 & 138 & 287 & \centering 12\% & \centering 22\%  & 38.04\\
            \bottomrule
        \end{tabular}
    \end{adjustbox}
    \caption{Dataset statistics for each variety, subset, split, and label type. \% Pos. Sent. and \% Pos. Sarc. indicates the proportion of samples with positive sentiment and true sarcasm respectively.}
    \label{tab:stats}
\end{table*}
\subsection{Annotation}\label{sec:annot}
We hire one annotator\footnote{Two of these are the authors of the paper who were also involved in the evaluation.} each for the three language varieties. The annotators assign the processed reviews and posts two labels: sentiment and sarcasm. The choice of labels given are \textit{negative}, \textit{positive}, and \textit{discard}. We instruct the annotators to use the \textit{discard} label for uninformative examples with no apparent polarity and in a few cases, for computer-generated messages. The texts with \textit{discard} label are discarded. 

We conduct an additional annotation exercise to evaluate the reliability of the sentiment and sarcasm annotations. For this, we randomly selected 50 instances from each variety, annotated by the original annotators, and hire a new independent annotator for each variety. The inter-annotator agreement between in the original and the independent annotators, measured by $\kappa$, are reported in the Table~\ref{tab:annot}. The absolute agreement values for sarcasm, though lower than those for sentiment, are comparable to prior work. In particular,~\citet{joshi-etal-2016-harnessing} report a $\kappa$ of 0.44 on their sarcasm labels. These scores suggest that both sentiment and sarcasm labels in our dataset are annotated with high degree of reliability. We compensate the annotators at the casual employment rate, equivalent to 22 USD per hour, as prescribed by the host institution. We provide the annotation guideline in Appendix~\ref{sec:annot-guide}.
%

\subsection{Dataset Statistics} The resultant \bench dataset contains annotations for two tasks: sentiment and sarcasm classification, for the two domains and three varieties. Table~\ref{tab:stats} provides a breakdown of the {annotated dataset (consisting of \google and \reddit)}, including the number of samples in each split (Train, Validation, and Test), along with average word length. The \% Pos. Sent. and \% Pos. Sarc. indicates the proportion of samples with positive sentiment and true sarcasm respectively. These values are the same for the train, validation, and test set since we perform stratified sampling. We present example comments along with their corresponding sentiment and sarcasm labels in Appendix~\ref{sec:dataset}. We also confirm that the text collected for Reddit subset of the \bench is generated in 2024. While this subset represents a static snapshot, its recency allows it to reflect contemporary linguistic usage effectively. 
\begin{figure*}[t!]
    \centering
    \includegraphics[width=\linewidth]{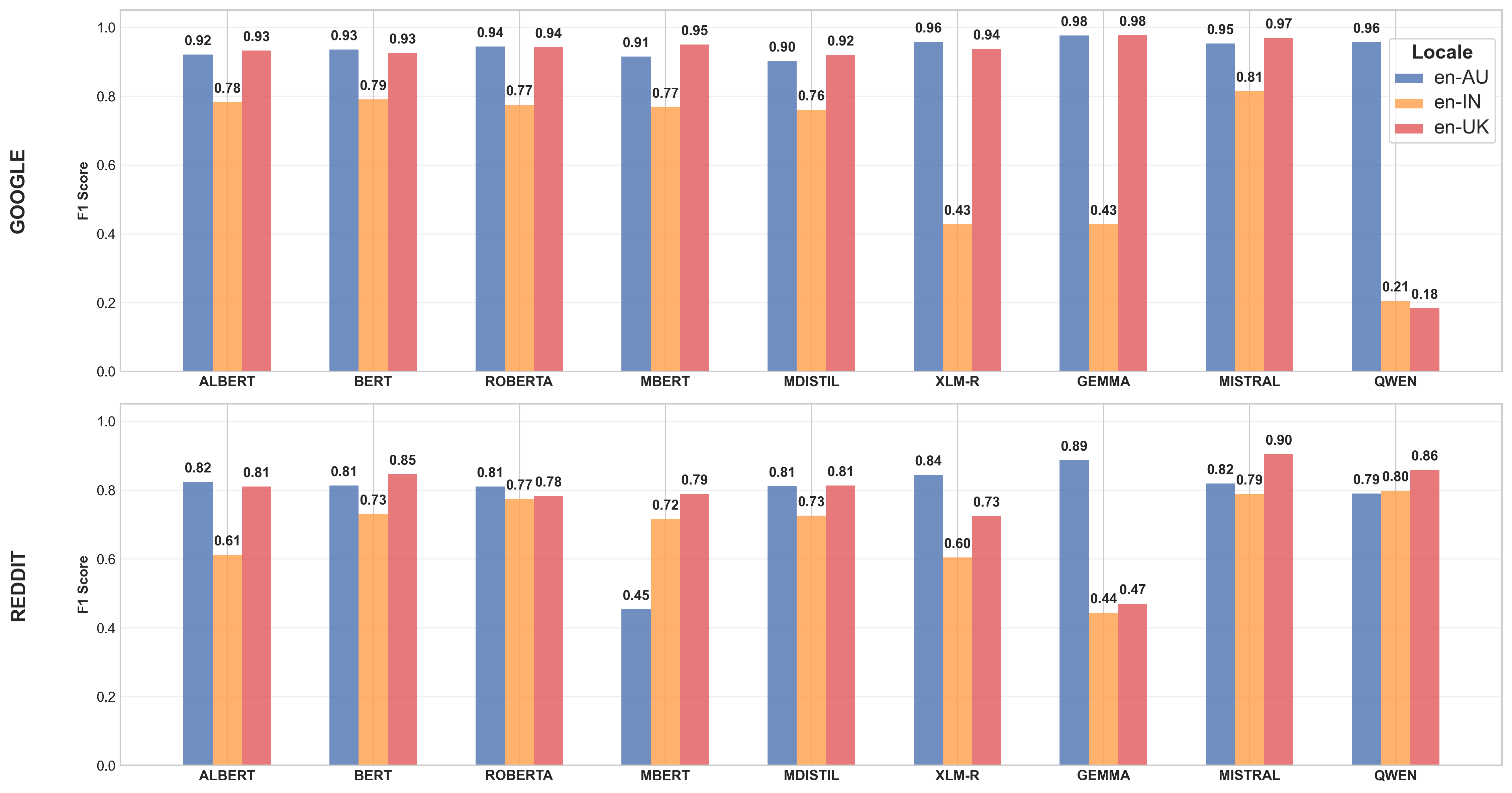}
    \caption{Performance comparison of various models on the sentiment classification task across different English varieties (en-AU, en-IN, and en-UK).}
    \label{fig:sent-all}
\end{figure*}
\begin{figure*}[t!]
    \centering
    \includegraphics[width=\linewidth]{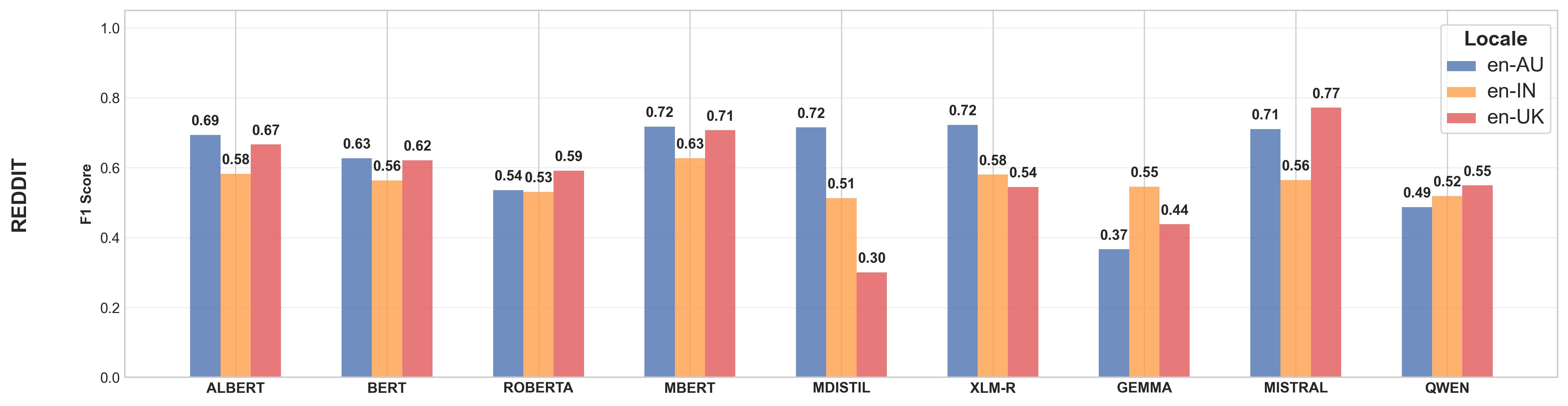}
    \caption{Performance comparison of various models on the sarcasm classification task across different English varieties (en-AU, en-IN, and en-UK).}
    \label{fig:sarc-all}
\end{figure*}

\section{Experiment Details}\label{sec:setup}
%
We conduct our experiments on a total of nine LLMs. These include {six encoder models, \textit{three}} pre-trained on English corpora: BERT-Large (\bert)~\cite{devlin-etal-2019-bert}, RoBERTa-Large (\roberta)~\cite{liu2019robertarobustlyoptimizedbert}, ALBERT-XXL-v2 (\albert)~\cite{lan2020albertlitebertselfsupervised}; and \textit{three} multilingual models: Multilingual-BERT-Base (\mbert), Multilingual-DistilBERT-Base (\mdis)~\cite{sanh2020distilbertdistilledversionbert}, XLM-RoBERTa-Large (\xlmr)~\cite{goyal-etal-2021-larger}. We also evaluate {three decoder models, \textit{one}} pre-trained on English corpora: Gemma2-27B-Instruct (\textsc{gemma})~\cite{gemmateam2024gemma2improvingopen} and \textit{two} multilingual models: Mistral-Small-Instruct-2409 (\textsc{mistral})~\cite{jiang2023mistral7b}, Qwen2.5-72B-Instruct (\textsc{qwen})~\cite{yang2024qwen2technicalreport}. The model-specific details including architecture, language coverage of pre-training corpus, and number of parameters are described in Appendix~\ref{sec:model_det}. 

While encoder models are fine-tuned with full precision, we fine-tune quantised decoder models using QLoRA~\cite{dettmers2023qloraefficientfinetuningquantized} adapters, targeting all linear layers. All models (including the variety predictor) are fined-tuned for 30 epochs, with a batch size of 8 and Adam optimiser. We choose an optimal learning rate by performing a grid search over the following values: 1e-5, 2e-5, and 3e-5. All experiments are performed using \textit{two} NVIDIA A100 80GB GPUs.

We fine-tune encoder models using cross-entropy loss, weighted on class distribution (\textit{positive}: `1' and \textit{negative}: `0' labels). In contrast, for decoder models, we perform zero-shot instruction fine-tuning~\cite{ouyang2022traininglanguagemodelsfollow}, using maximum likelihood estimation -- a standard learning objective for causal language modeling~\cite{jain-etal-2023-contraclm}. 

For decoder models, we use two task-specific prompts, one for each task. For sentiment classification, we the prompt the model with:
\begin{quote}
    \textit{``Generate the sentiment of the given text. 1 for positive sentiment, and 0 for negative sentiment. Do not give an explanation.''}
\end{quote}

Similarly, for sarcasm classification, we use: 
\begin{quote}
    \textit{``Predict if the given text is sarcastic. 1 if the text is sarcastic, and 0 if the text is not sarcastic. Do not give an explanation.''}
\end{quote}

Given the text and a task-specific prompt, the expected behaviour of the LLM is to generate either 1 (for \textit{positive}) or 0 (for \textit{negative}). The forms of these prompts are experimentally determined using a few test examples. Finally, we report our task performances on three macro-averaged \f, i.e., unweighted average, disregarding the imbalance in label distribution.
\section{Results}\label{sec:results}
We present our results and subsequent discussions to address the following questions: (a) How well do current LLMs perform on the benchmark tasks? (Section~\ref{sec:overall}); (b) How do factors such as model properties and domain affect the task performance? (Section~\ref{sec:factors}); (c) Can models trained on one language variety generalise well to others? (Section~\ref{sec:cross}).
\subsection{Task Performance}\label{sec:overall}
We first present our results on the two tasks where models are trained and evaluated on the same variety. Figures~\ref{fig:sent-all} and~\ref{fig:sarc-all} describe the model performances, reported using \f, on the sentiment and sarcasm classification tasks respectively. While sentiment classification is performed using both \google and \reddit subsets, due to the absence of sarcasm labels in \google subset (as shown in Table~\ref{tab:stats}), sarcasm classification is conducted only on \reddit subset. Table~\ref{tab:variety} shows that models, on average, perform better on both tasks when trained on en-AU variety, closely followed by performances on en-UK variety. Conversely, models consistently perform the worst on en-IN.
\begin{table}[t!]
    \centering
    \begin{tabular}{cccc}
    \toprule
    Domain-Task & en-AU & en-IN & en-UK\\\midrule[\heavyrulewidth]
    \google-Sentiment & 0.94 & 0.64 & 0.86\\
    \reddit-Sentiment & 0.78 & 0.69 & 0.78\\
    \reddit-Sarcasm & 0.62 & 0.56 & 0.58\\\hdashline
    $\mu$ & 0.78\high & 0.63\low & 0.74\\
    \bottomrule
    \end{tabular}
    \caption{Task performances, averaged across all models, on all varieties. $\mu$ represents the average performance across all domain-task pairs.}
    \label{tab:variety}
\end{table}
\paragraph{Sentiment Classification} 
Table~\ref{tab:arch_a} shows that, for sentiment classification, \mistral achieves the highest average performance across all varieties, with an \f of 0.91 on the \google subset and 0.84 on the \reddit subset. In contrast, \qwen reports the lowest average performance across all varieties on the \google subset, with an average \f of 0.45, while \gemma exhibits the lowest average performance across all varieties on the \reddit subset, with an \f of 0.60.

\paragraph{Sarcasm Classification} 
Similarly, for sarcasm classification, both \mistral and \mbert achieve the highest average performance across all varieties, with an \f of 0.68, as shown in Table~\ref{tab:arch_a}. In contrast, \gemma again reports the lowest average performance across all varieties, with an average \f of 0.45. The lower performance on sarcasm classification as compared to sentiment classification is expected because sarcasm often relies on not just subtle linguistic cues but also localised contemporary contextual information~\cite{abercrombie-hovy-2016-putting}. 

\subsection{Impact of Models and Domain}\label{sec:factors}
The average results for the models highlight the need to further probe into factors influencing model performances, namely, model properties and domain.

\paragraph{Model properties} The results ($\mu$ values) in the corresponding columns of  Table~\ref{tab:arch_a} describe the model performances (reporting \f), grouped based on model properties (encoder/decoder, mono/multilingual). The values show a consistent trend where encoder models report higher performance than decoder models across different tasks. This performance gap is expected as encoder architecture is inherently better suited for sequence classification tasks, while decoder architecture is optimised for text generation tasks. We also find that, although marginal, monolingual models report a higher average performance compared to multilingual models. This performance gap is prominent on the task involving \google subset, while multilingual models perform better on tasks involving \reddit subset.

\paragraph{\google vs. \reddit} Our findings (the last row $\mu$ for the task columns in Table~\ref{tab:arch_a}), indicate that models perform better on {\google subset}, achieving an \f of 0.81, compared to an \f of 0.75 on {\reddit subset}. This difference can be explained by the distinct writing styles of the two sources. Reviews in \google are generally more formal and informative, while posts and comments from \reddit often exhibit a short (as shown in Table~\ref{tab:stats}), conversational tone typical of social media. Additionally, posts and comments in the \reddit subset come from forums frequented by local speakers, which means the language and expressions used are more reflective of the variety and cultural nuances. We present results from cross-domain evaluation in the Appendix~\ref{sec:add-res}.

\begin{table*}[t!]
    \begin{adjustbox}{width=0.85\linewidth, center}
        \begin{tabular}{cccccc}
        \toprule
        Model & Enco. & Mono. & \google-Sentiment & \reddit-Sentiment & \reddit-Sarcasm\\\midrule[\heavyrulewidth]
        \albert & \checkmark & \checkmark & 0.88 & 0.75 & 0.65\\
        \bert & \checkmark & \checkmark & 0.88 & 0.80 & 0.60\\
        \roberta & \checkmark & \checkmark & 0.89 & 0.79 & 0.55\\
        \mbert & \checkmark & \xmark & 0.88 & 0.65 & 0.68\\
        \mdis & \checkmark & \xmark & 0.86 & 0.78 & 0.51\\
        \xlmr & \checkmark & \xmark & 0.77 & 0.73 & 0.62\\
        \gemma & \xmark & \checkmark & 0.79 & 0.60 & 0.45\\
        \mistral & \xmark & \xmark & 0.91 & 0.84 & 0.68\\
        \qwen & \xmark & \xmark & 0.45 & 0.82 & 0.52\\\hdashline
        \centering\multirow{2}{2em}{$\mu$} & \checkmark: 0.74 & \checkmark: 0.72 & \multirow{2}{2em}{0.81\high} & \multirow{2}{2em}{0.75} & \multirow{2}{2em}{0.59\low}\\\cmidrule(lr){2-2}\cmidrule(lr){3-3}
        & \xmark: 0.67 & \xmark: 0.71 & & & \\
        \bottomrule
        \end{tabular}  
    \end{adjustbox}
        \caption{Model performances, averaged across all varieties, on sentiment and sarcasm classification tasks across two subsets (\google and \reddit). $\mu$ represents the average performance across all models. `Enco.' represents encoder models. `Mono.' represents monolingual models.}
        \label{tab:arch_a}
\end{table*}
\begin{figure*}[t!]
    \centering
    \includegraphics[width=0.7\linewidth]{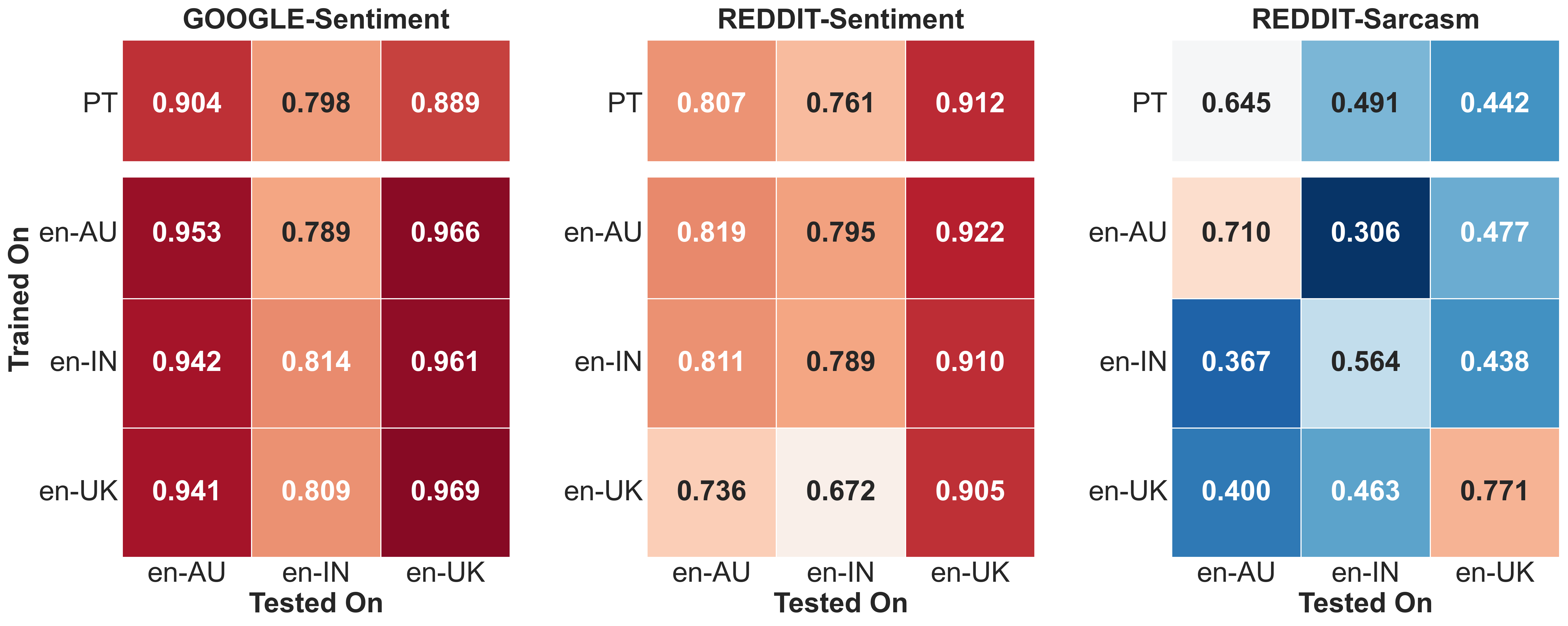}
    \caption{Cross-variety performance analysis of \mistral. The figure compares three different scenarios: pre-trained (PT), in-variety fine-tuning, and cross-variety fine-tuning for sentiment and sarcasm classification across all varieties.}
    \label{fig:cross-mistral}
\end{figure*}
\subsection{Cross-variety Evaluation}\label{sec:cross}
Finally, we perform cross-variety evaluation. This refers to the scenario where the datasets used to train and test a model are from different groups (i.e., trained on en-AU, tested on en-IN, for example).

We report the cross-variety results of our best-performing model, \mistral. Figure~\ref{fig:cross-mistral} compares the performance of \mistral across three variants: pre-trained, in-variety fine-tuning, and cross-variety fine-tuning, for the three dataset subsets using colour-coded matrices. The rows indicate the subset that the model is trained on (i.e., en-AU, en-IN, and en-UK), along with PT, indicating that the pre-trained \mistral is used. Similarly, the columns indicate the test subset (i.e., en-AU, en-IN, and en-UK). The pre-trained variant of the model (the first row in each matrix) achieves high \f for en-AU and en-UK, for the sentiment classification task. Fine-tuning the model improves in-variety performance (shown by the principal diagonal of the matrices), with noticeable gains in \f on both subsets. The model performance on cross-variety evaluation remains relatively stable across different varieties of English, indicating that domain-specific data plays a minor role in influencing cross-variety generalisation for sentiment classification.

The model performs significantly worse on sarcasm classification compared to sentiment classification. While in-variety evaluation shows that fine-tuning improves over the pre-trained model with substantial increases in \f, it negatively impacts cross-variety generalisation. This means there is a notable decrease in \f across other varieties when the model is fine-tuned on a specific variety. This supports the idea that sarcasm classification requires the model to grasp linguistic and cultural nuances specific to a variety, making generalisation across varieties challenging.

\begin{table}[t!]
    \begin{adjustbox}{width=\linewidth,center}
        \begin{tabular}{cccccc}
        \toprule
            Variety & Sample & \textsc{dial} & \textsc{coll} & \textsc{cont} & \textsc{code}\\\midrule[\heavyrulewidth]
            en-AU & 70 & 9 & 28 & 6 & -\\
            en-IN & 90 & 97 & 33 & 3 & 8\\
            en-UK & 53 & 7 & 15 & 5 & - \\
            \bottomrule
        \end{tabular}
    \end{adjustbox}
    \caption{Counts of identified features in misclassified examples from \mistral for each variety. \textsc{dial} denotes dialect features, \textsc{coll} denotes locale-specific colloquial expressions, \textsc{cont} denotes instances requiring additional context, and \textsc{code} denotes occurrences of code-mixed text.}
    \label{tab:err-anal}
\end{table}
\section{Error Analysis}\label{sec:err-anal}
We randomly sample up to 30 examples misclassified by \mistral, from each variety and domain-task configuration. We then manually analyse these misclassified examples to identify pervasive or obligatory dialect features, as defined by eWAVE~\cite{ewave}. A dialect feature is defined as any lexical or syntactic variation from standard English. In addition, we identify other features, including locale-specific colloquial expressions, instances that require additional context, and occurrences of code-mixed text. Table~\ref{tab:err-anal} summarises the sample sizes and the counts of these identified features for each variety. Detailed examples for each feature type are provided in Appendix~\ref{sec:det-err}.

The en-IN variety exhibits a high count of dialect features (97) relative to its sample size (90), suggesting that regional variations significantly challenge model performance. Locale-specific colloquial expressions are prevalent in all varieties (en-AU: 28; en-IN: 33; en-UK: 15), highlighting the impact of regional lexicons on model performance. These findings motivate future research in sentiment and sarcasm classification to better accommodate such colloquial expressions for inner-circle varieties, and to additionally adapt to dialect features for outer-circle varieties of English.
\section{Related work}\label{sec:relwork}
NLP Benchmarks such as GLUE \cite{wang-etal-2018-glue}, SuperGLUE \cite{NEURIPS2019_4496bf24}, and DynaBench \cite{kiela-etal-2021-dynabench} have played a pivotal role in evaluating LLMs on multiple NLP tasks, including sentiment classification. However, these benchmarks are limited in their ability to capture the linguistic nuances of non-standard language varieties. Efforts to address this include datasets for African languages and dialects \cite{muhammad2023afrisentitwittersentimentanalysis}, Arabic dialects \cite{elmadany-etal-2023-orca}, and CreoleVal \cite{https://doi.org/10.48550/arxiv.2310.19567} which provides datasets for 28 creole languages. More recent progress includes DialectBench~\cite{faisal-etal-2024-dialectbench}, which presents a collection of dialectal datasets and associated NLP models across 281 dialectal varieties for several tasks. However, the benchmark does not contain English dialectal datasets for sentiment and sarcasm classification.

%
Similarly, several sarcasm classification datasets for standard variety of English exist, consisting of reddit comments \cite{khodak-etal-2018-large},  amazon reviews \cite{filatova-2012-irony}, and tweets \cite{ptacek-etal-2014-sarcasm, painter-etal-2022-utilizing, abercrombie-hovy-2016-putting, oprea-magdy-2020-isarcasm}. While datasets like ArSarcasm-v2 \cite{abu-farha-etal-2021-overview} present annotations for sarcasm, sentiment, and the dialect of each tweet for Arabic, there is a notable absence of sarcasm classification datasets that account for non-standard varieties of English.  Similar to their work, we also create a dataset that contains both sentiment and sarcasm labels. To the best of our knowledge, \bench is the {first benchmark for sentiment and sarcasm classification specifically for varieties of English}.

\section{Conclusion}\label{sec:concl}
This paper presents \bench, a benchmark for sentiment and sarcasm classification across three varieties of English: Australian (en-AU), Indian (en-IN), and British (en-UK). Our evaluation spans nine LLMs, six encoders and three decoders, assessed on datasets collected from Google Places reviews (\google) and Reddit comments (\reddit). The models perform consistently better on en-AU and en-UK (i.e., inner-circle varieties) than on en-IN (i.e., the outer-circle variety) for both tasks. We verify the variety representation of collected data using: (a) manual annotation and (b) automated validation. We also perform an additional annotation exercise to verify the reliability of the annotations. Although models report high performance on sentiment classification (\f of 0.81 and 0.75, for \google and \reddit respectively), they struggle with detecting sarcasm (\f of 0.59 on \reddit), indicating that sarcasm classification is still largely unsolved. This is supported by our cross-variety evaluation which reveals limited generalisation capability for sarcasm classification, where cultural and contextual understanding is crucial. Notably, monolingual models marginally outperform multilingual models, suggesting that language diversity in pre-training does not extend to varieties of a language. Our error analysis provides potential considerations to drive future research in sentiment and sarcasm classification to better accommodate colloquial expressions for inner-circle varieties, and dialect features for outer-circle varieties of English. In conclusion, \bench provides a resource to measure bias in LLMs towards non-standard English varieties for the two tasks.


\section*{Limitations}
Our assumption of national varieties as representative forms of dialect is a simplification. Within each locale, there are significant regional, sociolectal, and generational differences. For example, Australian English spoken in Sydney may differ significantly from that in Perth. The language used in online communications evolves rapidly, driven by cultural trends, viral content, and changing societal norms. Phrases, slang, and expressions can gain popularity and fade quickly. A static dataset, such as ours, is hence limited in reflecting these continuous and dynamic changes in language use. While our efforts to remove non-English text were thorough, there may still be instances where language mixing or code-switching occurs, especially in the Indian English subset, affecting model performance. The annotation process was limited by the number of human annotators and their subjective interpretations. While we perform an additional annotation exercise to validate the annotation process, use of a single annotator can introduce individual biases. We also note that comments from Reddit may differ from other social media sites, such as Twitter/X, in terms of user demographics and discourse style. This potentially influences the nature of topic-based discussions captured in our dataset. Our reliance on Reddit data is due to cost-related constraints associated with accessing Twitter/X API.



\section*{Ethical Considerations}
The project received ethics approval from the Human Research Ethics Committee at UNSW, Sydney (reference number: iRECS6514), the host organisation for this research.
We treat the reviews as separate texts and do not aggregate them by user in any way. The Reddit comments and Google reviews used in this paper are from publicly available posts, accessible through official APIs. We adhere to the terms of service and policies outlined by these platforms while gathering data. We do not attempt to de-identify any users or collect aggregate information about individual users. The dataset, including sentiment and sarcasm annotations, is shared in a manner that preserves user privacy and abides by the rules and guidelines of the source platforms.

\section*{Acknowledgements}
This research was funded by Google's Research Scholar grant, awarded to Aditya Joshi and Diptesh Kanojia in 2024. 


\bibliography{custom}
\onecolumn
\appendix
\section{Annotation Guidelines}\label{sec:annot-guide}
\subsection*{Task Overview}  Your task involves determining whether each Google review or Reddit comment expresses positive or negative sentiment and whether it uses sarcasm. 
\subsection*{Annotation Instructions}%
For Sentiment Classification:
\begin{itemize}
    \item Positive Sentiment (Label 1): Annotate a comment as positive if it reflects favorable emotions such as happiness, satisfaction, agreement, or excitement.  
    \item Negative Sentiment (Label 0): Annotate a comment as negative if it conveys unfavorable emotions such as sadness, disappointment, frustration, or criticism. 
    \item Neither Positive nor Negative (Discarded, Label 2): discard the comment if it does not convey clear positive or negative sentiment (mark the same sarcasm label as 2). 
\end{itemize}
For Sarcasm Classification:
\begin{itemize}
\item Sarcastic Comment (Label 1): Annotate if the comment uses irony or mockery to express contempt or ridicule.  
\item Non-Sarcastic Comment (Label 0): Annotate if the comment is straightforward and does not employ sarcasm or irony. 
\end{itemize}
\subsection*{Additional Guidelines} 
Consider the tone and context of the comment when assigning sentiment or sarcasm labels. 
For sarcasm, look for indicators such as exaggerated language, contradictions, or unexpected statements. 
Base your decision on the sentiment explicitly expressed in the text, avoiding personal biases. 
It is acceptable to mark a comment as undecided or seek clarification if you cannot confidently assign a sentiment label. 
\section{Dataset Examples}\label{sec:dataset}

\begin{table*}[h!]
\centering
        \begin{tabular}{cccc}
            \toprule
            Variety & \centering Comment & Sentiment & Sarcasm \\\midrule[\heavyrulewidth]
            en-AU  & Well set out, good stock and very friendly country staff & 1 & 0 \\
            en-AU  & S**t. Christ isn’t back, is he? & 0 & 1 \\\midrule[\heavyrulewidth]
            en-IN  & Good quality foods are supplied here & 1 & 0 \\
            en-IN  & Coz we all have free internet. & 0 & 1 \\\midrule[\heavyrulewidth]
            en-UK  & Traditional friendly pub. Excellent beer & 1 & 0 \\
            en-UK  & What a brave potatriot & 0 & 1 \\ \bottomrule
        \end{tabular}

    \caption{Reviews and comments from different varieties with their corresponding annotated sentiment and sarcasm labels.}
    \label{tab:examples}
\end{table*}
The examples presented in Table~\ref{tab:examples} illustrate the diversity of language use and the challenges involved in detecting both sentiment and sarcasm across the varieties. 

\section{Additional Model Details}\label{sec:model_det}
We describe additional model details, including architecture, the language coverage in their pre-training data, and the number of trainable parameters in Table~\ref{tab:model_det}.
\begin{table}[h!]
    \centering
    \begin{tabular}{cccc}
    \toprule
    Model & Arch. & Lang. & \# Params\\\midrule[\heavyrulewidth]
    \albert & Encoder & English & 223 M\\
    \bert & Encoder & English & 340 M\\
    \roberta & Encoder & English & 355 M\\
    \mbert & Encoder & Multilingual & 177 M\\
    \mdis & Encoder & Multilingual & 134 M\\
    \xlmr & Encoder & Multilingual & 355 M\\\midrule[\heavyrulewidth]
    \gemma & Decoder & English & 27 B\\
    \mistral & Decoder & Multilingual & 22 B\\
    \qwen & Decoder & Multilingual & 72 B\\
    \bottomrule
    \end{tabular}
    \caption{Details of models including architecture (Arch.), language coverage in pre-training data (Lang.), and number of trainable parameters (\# Params). Here, M is millions and B is billion.}
    \label{tab:model_det}
\end{table} 
\section{Cross-domain Evaluation}\label{sec:add-res}
\begin{figure*}[h!]
    \centering
    \includegraphics[width=\linewidth]{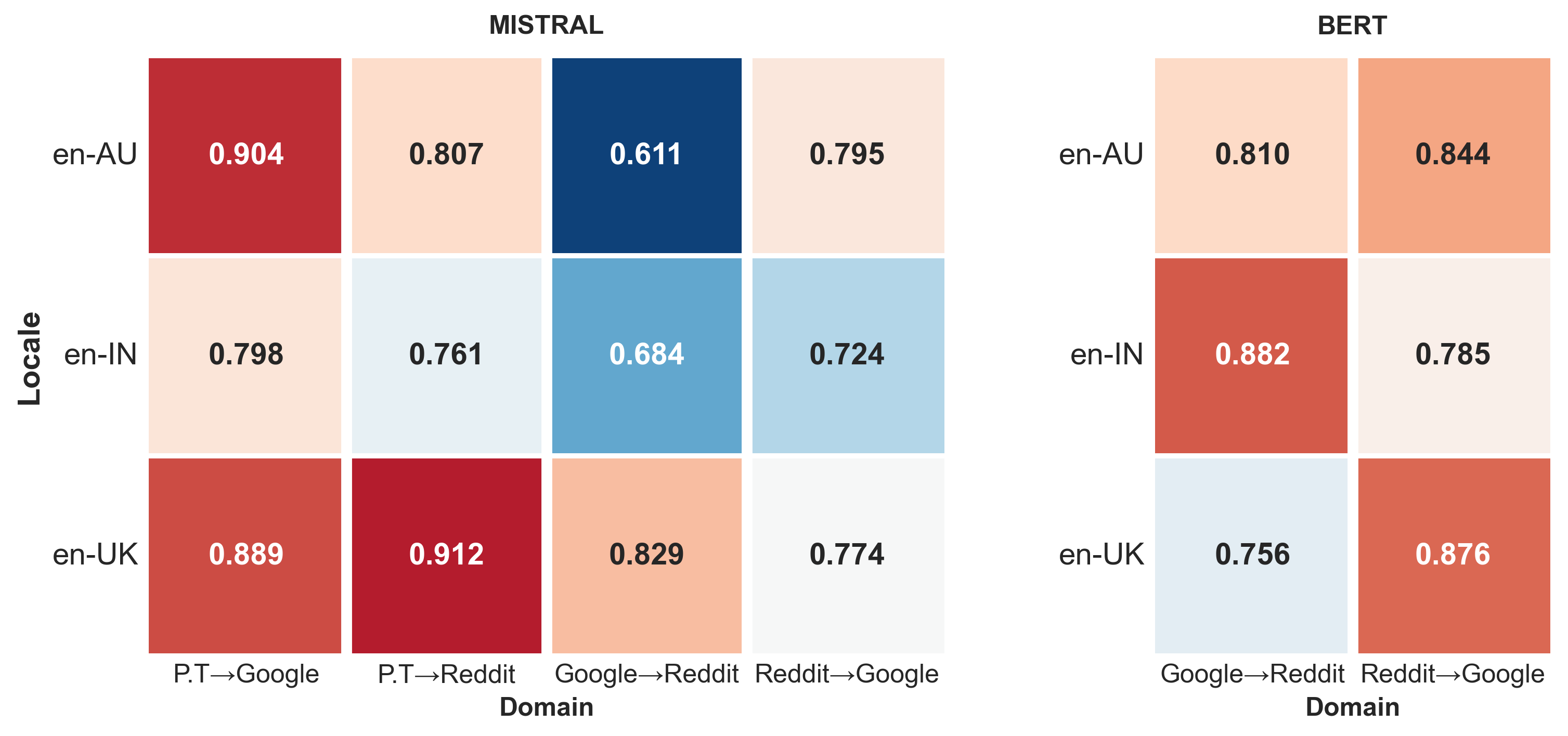}
    \caption{Cross-domain performance analysis of \mistral and \bert comparing a decoder and an encoder model.}
    \label{fig:cross-domain}
\end{figure*}

We conduct experiments with domain data from \google, and \reddit, to assess the cross-domain robustness of our sentiment classification models. Models trained on one domain are tested on the other domain, and vice-versa, for each of the three varieties of English. Figure \ref{fig:cross-domain} shows that models perform better when trained on in-domain data, for both \mistral and \bert. We observe that pre-trained \mistral performs better for all varieties in comparison to results from the cross-domain experiments. 
\begin{figure*}[h!]
    \centering
    \includegraphics[width=\linewidth]{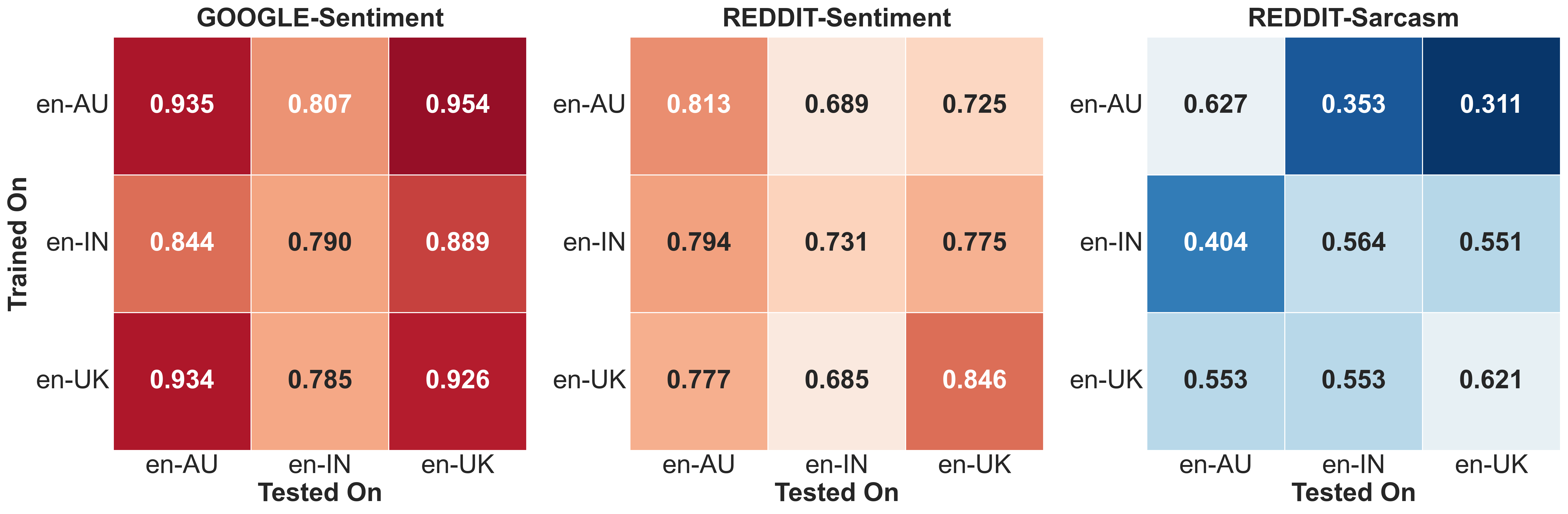}
    \caption{Cross-variety performance analysis of \bert. The figure compares two different scenarios: in-variety fine-tuning, and cross-variety fine-tuning for sentiment and sarcasm classification across all varieties.}
    \label{fig:cross-bert}
\end{figure*}
The corresponding results for \bert are in Figure~\ref{fig:cross-bert}.

While \mistral outperforms \bert at both in- and cross-domain experiments, there is significant variation in cross-domain results, highlighting the challenge in \textit{generalization}, given domain variance in fine-tuned data. Moreover, the direction of cross-domain experiments also affects performance. Both models perform better for \reddit $\rightarrow$ \google, suggesting that \textit{models may be better at transferring sentiment from a relatively informal domain}. The overall difference in cross-domain performance is significantly higher for inner-circle varieties, compared to en-IN, indicating that fine-tuning with mixed-domain data may be less effective for such varieties. Across the varieties, however, en-UK, an inner-circle variety, performs better for \google $\rightarrow$ \reddit, whereas for the \bert model, this is observed with the en-IN variety.
\section{Detailed Feature Types}\label{sec:det-err}
We present detailed feature types identified in misclassified samples from different English varieties using the \mistral model. Each table lists the feature, an example illustrating the feature, and the frequency count. Table~\ref{tab:dial-au} shows features found in en-AU samples, Table~\ref{tab:dial-ind} shows features found in en-IN samples, and Table~\ref{tab:dial-uk} shows features found in en-UK samples.
\begin{table*}[h!]
    \begin{adjustbox}{width=\linewidth,center}
        \begin{tabular}{p{6.6cm}p{8.1cm}c}
        \toprule
            Feature & Example & Count\\\midrule[\heavyrulewidth]
            \textsc{{dial}} & & \\\cmidrule(lr){1-1}
            \centering\textsc{levelling of past tense verb forms} & \textit{when we \underline{order} passionfruit cheecake and never \underline{get} it} & 2\\
            \centering\textsc{pronoun drop} & \textit{(We) also ordered 2 noodles, fish and dessert} & 3\\
            \centering\textsc{there's with plural subjects} & \textit{\underline{There's} been plenty of fossils and opals uncovered} & 1\\
            \centering\textsc{no auxiliary in yes/no questions} & \textit{(Do) people still watch commercial TV?} & 2\\\midrule
            \textsc{coll} & \textit{Hi \underline{Hospo} workers afternoon when you wake up} & 28\\
            \textsc{cont} & \textit{Beige night} & 6\\
            \bottomrule
        \end{tabular}
    \end{adjustbox}
    \caption{Features identified in the en-AU samples misclassified by \mistral.}
    \label{tab:dial-au}
\end{table*}

\begin{table*}[h!]
    \begin{adjustbox}{width=\linewidth,center}
        \begin{tabular}{p{4.5cm}p{10cm}c}
        \toprule
            Feature & Example & Count\\\midrule[\heavyrulewidth]
            \textsc{{dial}} & & \\\cmidrule(lr){1-1}
            \centering \textsc{article omission} & \textit{Kamal is playing (a) cameo role} & 35\\
            \centering \textsc{pronoun drop} & \textit{(I) can't wait for animal park man} & 22\\
            \centering \textsc{object/subject fronting} & \textit{\underline{Laws of India doesn't apply on white man}, he is above everything} & 12\\
            \centering \textsc{‘very’ as qualifier} & \textit{quantity is \underline{very very} less} & 10\\
            \centering \textsc{copula omission} & \textit{(I \underline{am}) Waiting for cash to arrive in a tempo} & 18\\\midrule
            \textsc{coll} & \textit{It means \underline{benstokes}} & 33\\
            \textsc{cont} & \textit{Alia Advani as her earlier name suits better.} & 3\\
            \textsc{code} & \textit{Bridge \underline{chori ho jaaega}} & 8\\
            \bottomrule
        \end{tabular}
    \end{adjustbox}
    \caption{Features identified in the en-IN samples misclassified by \mistral.}
    \label{tab:dial-ind}
\end{table*}

\begin{table*}[t!]
    \begin{adjustbox}{width=\linewidth,center}
        \begin{tabular}{p{6.6cm}p{8.1cm}c}
        \toprule
            Feature & Example & Count\\\midrule[\heavyrulewidth]
            \textsc{{dial}} & & \\\cmidrule(lr){1-1}
            \centering\textsc{was for conditional were} & \textit{if it \underline{was} in the autumn.} & 1\\
            \centering\textsc{group genitive} & \textit{\underline{their and their} old boy network} & 1\\
            \centering\textsc{levelling of past tense verb forms} & \textit{but \underline{could been} more spicy.} & 1\\
            \centering\textsc{pronoun drop} & \textit{(It) took just 2 and a half minutes to lose his head} & 4\\\midrule
            \textsc{coll} & \textit{\underline{Warra} calm and coherent interview.} & 15\\
            \textsc{cont} & \textit{Haha yup!} & 5\\
            \bottomrule
        \end{tabular}
    \end{adjustbox}
    \caption{Features identified in the en-UK samples misclassified by \mistral.}
    \label{tab:dial-uk}
\end{table*}

\end{document}